\documentclass[10pt,twocolumn,letterpaper]{article}

\usepackage{iccv}
\usepackage{times}
\usepackage{epsfig}
\usepackage{graphicx}
\usepackage{amsmath}
\usepackage{amssymb}

\usepackage{graphicx}
\usepackage{amsmath}
\usepackage{amssymb}
\usepackage{booktabs}
\usepackage{times}
\usepackage{epsfig}
\usepackage{enumitem}

\usepackage{multirow}
\usepackage{soul}
\usepackage{pifont} 
\usepackage[T1]{fontenc}
\usepackage{mathabx,graphicx}
\usepackage{makecell}
\usepackage[dvipsnames]{xcolor}
\usepackage{color, colortbl}
\usepackage{caption}

\def\CircleArrowright{\ensuremath{%
  \rotatebox[origin=c]{310}{$\circlearrowright$}}}

\newcommand{\xmark}{\ding{55}} %
\newcommand\Tstrut{\rule{0pt}{2.3ex}}         
\newcommand{\vlnbert}{VLN$\protect\CircleArrowright$BERT}

\usepackage[pagebackref=true,breaklinks=true,letterpaper=true,colorlinks,bookmarks=false]{hyperref}

\iccvfinalcopy 

\def\iccvPaperID{9647} 

\begin{document}

\title{Learning Navigational Visual Representations with Semantic Map Supervision}

\author{Yicong Hong$^{12,{\dag}}$ \quad Yang Zhou$^1$ \quad Ruiyi Zhang$^1$\\
Franck Dernoncourt$^1$ \quad Trung Bui$^1$ \quad
Stephen Gould$^2$ \quad Hao Tan$^1$ \\
$^1$Adobe Research \quad $^2$The Australian National University \\
{\tt\small mr.yiconghong@gmail.com} \\
{\tt\small stephen.gould@anu.edu.au, \{yazhou,ruizhang,dernonco,bui,hatan\}@adobe.com} \\
{\tt\small Project URL: \url{https://github.com/YicongHong/Ego2Map-NaViT}}
}

\maketitle

\begin{abstract}
   Being able to perceive the semantics and the spatial structure of the environment is essential for visual navigation of a household robot. However, most existing works only employ visual backbones pre-trained either with independent images for classification or with self-supervised learning methods to adapt to the indoor navigation domain, neglecting the spatial relationships that are essential to the learning of navigation. Inspired by the behavior that humans naturally build semantically and spatially meaningful cognitive maps in their brains during navigation, in this paper, we propose a novel navigational-specific visual representation learning method by contrasting the agent's egocentric views and semantic maps (Ego$^2$-Map). We apply the visual transformer as the backbone encoder and train the model with data collected from the large-scale Habitat-Matterport3D environments. Ego$^2$-Map learning transfers the compact and rich information from a map, such as objects, structure and transition, to the agent's egocentric representations for navigation. Experiments show that agents using our learned representations on object-goal navigation outperform recent visual pre-training methods. Moreover, our representations significantly improve vision-and-language navigation in continuous environments for both high-level and low-level action spaces, achieving new state-of-the-art results of 47\% SR and 41\% SPL on the test server.
   {\let\thefootnote\relax\footnote{{$^{\dag}$Intern at Adobe Research.}}}
\end{abstract}


\section{Introduction}
\label{sec:introduction}

Visual representations for navigation should capture the rich semantics and the complex spatial relationships of the observations, which helps the agent to recognize visual entities and its transition in space for effective exploration. However, previous works usually adopt visual backbones which only focus on capturing the semantics of a static image, ignoring its connection to the agent or the correspondence to other views in a continuous environment~\cite{anderson2018vln,chaplot2019activeslam,hong2020language,hong2020recbert,krantz2020vlnce,pashevich2021episodic,ye2021auxiliary}. One common approach is to apply encoders pre-trained for object/scene classification (\textit{e.g.} ResNet~\cite{he2016resnet} on ImageNet~\cite{russakovsky2015imagenet}/Places365~\cite{zhou2017places365}), object detection (\textit{e.g.} RCNN~\cite{he2017maskrcnn,ren2015faster} on VisualGenome~\cite{krishna2017genome}/MS-COCO~\cite{lin2014mscoco}) or semantic segmentation (\textit{e.g.} RedNet~\cite{jiang2018rednet} on SUN RGBD~\cite{song2015sunrgbd}), or more recently, to use CLIP~\cite{radford2021clip}, which is trained for aligning millions of images and texts to encode agent's RGB observations~\cite{gadre2022onwheels,khandelwal2022embclip,shen2021benefit}. Despite the increasing generalization ability of the features and rising zero-shot performance on novel targets, there still exists a large visual domain gap between these features and the features suitable for navigation since they lack the expressiveness of spatial relationships. For example, the connection between time-dependent observations and the correspondence between egocentric views and the spatial structure of an environment, which are important to decision making during navigation.


We suggest that there are two main difficulties in learning navigational-specific visual representations in previous research; First, there lacked a large-scale and realistic dataset of indoor environments. Popular datasets such as Matterport3D~\cite{chang2017matterport3d} and Gibson~\cite{xia2018gibson} provide traversable scenes rendered from real photos, but either the number of scenes is very limited, or the quality of the 3D scan is low. Synthetic datasets such as ProcTHOR~\cite{deitke2022procthor} contain 10K generated scenes, but they are unrealistic and simple and do not capture the full complexity of real-world environments. Second, fine-tuning visual encoders while learning to navigate is very expensive because of the long traveling horizon, especially in tasks that require extensive exploration~\cite{batra2020objectnav,anderson2020rxr,mezghani2021imagenav,shridhar2020alfred}. Moreover, due to the scene scarcity, fine-tuning visual encoders is unlikely to generalize well to the novel environments. 
To address these problems, a large amount of research has been dedicated to augmenting the environments, either editing the agent's observations \cite{li2022envedit,tan2019envdrop} or adding objects~\cite{maksymets2021thda}, building new spatial structures~\cite{liu2021envmixup}, applying web-images to replace the views of the same categories~\cite{guhur2021airbert} or generating new scenes~\cite{koh2022simple,koh2021pathdreamer,li2023panogen}. 
Unlike augmentation, which essentially requires the agent's policy network to adapt new visual data, another strategy is to first pre-train the visual encoder before learning to navigate. Those methods usually apply images captured in 3D scenes and perform self-supervised learning or tune the encoder with proxy tasks (\eg masked region modeling, angular prediction) to mitigate the domain gap and introduce spatial awareness~\cite{chen2021hamt,chen2022duet,yadav2022ovrl,ye2021auxiliary}. 
Furthermore, recent works tend to apply CLIP~\cite{radford2021clip} to process visual inputs, as the model can provide powerful representations for grounding semantic and spatial concepts, hence facilitating learning~\cite{an20221st,gadre2022onwheels,khandelwal2022embclip,shah2022lmnav,shen2021benefit,shridhar2022cliport}. 
Although these methods have demonstrated significant improvements on navigation performance, they either only use more data without addressing the learning problem, not generalizable across different tasks, or ignore the spatial correspondence between consecutive frames which the agent will capture as it moves.

Inspire by the research in cognitive science that humans naturally build virtual maps from perspective observations that are helpful to track semantics, spaces and movements during navigation~\cite{epstein2017cognitive,tolman1948cognitive,wang2002human}; we propose a contrastive learning method between the agent's \textbf{Ego}centric view \textbf{pairs} and top-down semantic \textbf{Maps} (Ego$^2$-Map) for training visual encoders appropriate for navigation. Specifically, we first sample RGBD images and create corresponding semantic maps~\cite{cartillier2021semmap} from the large-scale Habitat-Matterport3D environments (HM3D)~\cite{ramakrishnan2021hm3d,savva2019habitat} which contains hundreds of photo-realistic scenes, obtaining abundant and effective visual data. Then, we encode the sampled RGBD observations and the semantic maps with two separate visual backbones, respectively, and train the entire model with our Ego$^2$-Map contrastive learning. Once trained, the RGBD encoder will be plugged into an agent's network to facilitate visual perception. We argue that the map contains very rich and compact visual clues such as spatial structure, accessible areas and unexplored regions, object entities and their arrangement, as well as the agent's transition in the environment that are essential to navigation. Compared to existing online mapping-based approaches~\cite{chaplot2020object,chen2022weakly,georgakis2021learning,irshad2021sasra}, Ego$^2$-Map learning efficiently produces visual features that imply a complete map of the space known a view priori, while the online map is only a partial map. Importantly, Ego$^2$-Map explores a new possibility of modeling semantics and structures from simple RGBD inputs, which its resulting features are directly applicable to non-mapping-based navigational models and effectively generalizable to different tasks.

We mainly evaluate the features learned from Ego$^2$-Map on the Room-to-Room Vision-and-Language Navigation in Continuous Environments  (R2R-CE)~\cite{anderson2018vln,krantz2020vlnce} task, which requires an agent to navigate in photo-realistic environments following human natural language instructions such as ``\textit{Leave the bedroom and enter the kitchen. Walk forward and take a left at the couch. Stop in front of the window}.'' Addressing R2R-CE highly relies on exploiting the correspondence between contextual clues and the agent's observations, hence it is crucial for the visual encoder to provide semantic and structural meaningful representations. We found that the proposed Ego$^2$-Map features significantly boost the agent's performance, obtaining +3.56\% and +5.10\% absolute SPL improvements over the CLIP baseline~\cite{radford2021clip} under the settings of high-level and low-level action spaces, respectively, and achieves the new best results on the R2R-CE test server. Moreover, our experiments show that Ego$^2$-Map learning also outperforms other visual representation learning methods such as OVRL~\cite{yadav2022ovrl} on the Object-Goal Navigation task (ObjNav), suggesting the strong generalization potential of the proposed methods.

\section{Related Work}
\label{sec:related_works}

\paragraph{Visual Navigation}

A great variety of scenarios have been proposed to learn visual navigation in photo-realistic environments~\cite{chang2017matterport3d,deitke2020robothor,li2022igibson,savva2019habitat,shridhar2020alfred,szot2021habitat2,xia2018gibson} with different modalities of inputs, targets and action spaces~\cite{anderson2018vln,batra2020objectnav,chaplot2020neural,chen2020soundspaces,krantz2020vlnce,padmakumar2022teach,qi2020reverie,wijmans2019ddppo}. Due to the distinct nature of the tasks, diverse methods are investigated accordingly. For instance, in Object-Nav, which only provides a high-level goal and requires exploration, mapping-based methods are frequently applied~\cite{chaplot2020object,georgakis2021learning,luo2022stubborn,ramakrishnan2022poni,zhu2022navigating}, whereas in vision-and-language navigation (VLN)~\cite{anderson2018vln}, large vision-language models are employed to match instructions and agent's observations~\cite{hong2020recbert,majumdar2020improving,qiao2022hop,shah2022lmnav,zhou2023navgpt} to perform panoramic actions~\cite{fried2018speaker}. Recently, there is an emerging trend of scaling up the training data to address the common data scarcity issue, either in terms of the number of environments~\cite{deitke2022procthor,ramakrishnan2021hm3d} or the amount of supervision~\cite{chen2022learning,ramrakhya2022habweb,wang2023scalevln}. Unlike most of the previous works, we focus on improving the visual backbone, aiming to produce robust and generalizable visual representations specialized for navigation.

\begin{figure*}[t]
  \centering
  \includegraphics[width=\textwidth]{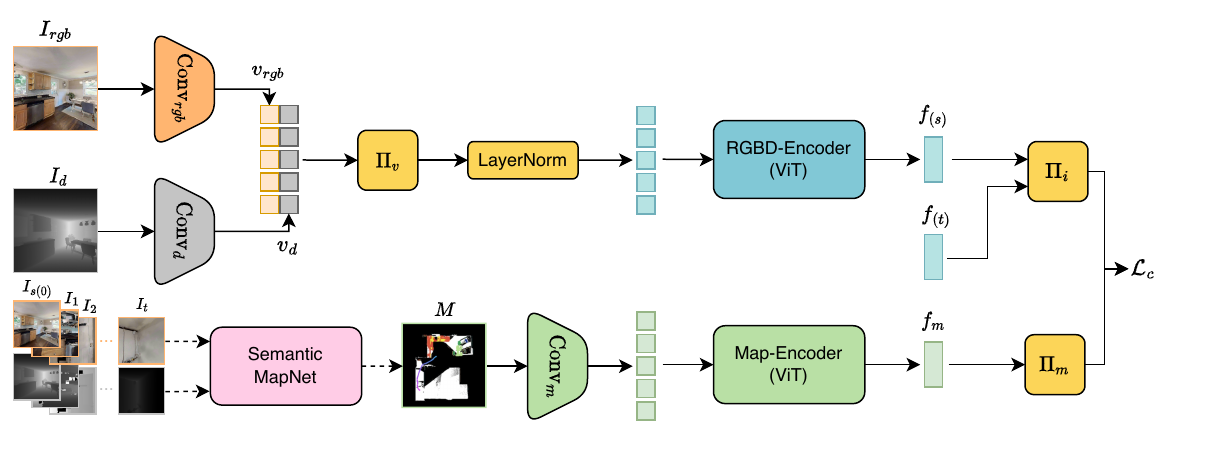}
  \caption{Network architecture. The Semantic MapNet~\cite{cartillier2021semmap} is applied to draw a semantic map using observations captured from a source to a target position. Then, two learnable visual encoders are applied to encode the egocentric views and the map, respectively, followed by non-linear headers $\boldsymbol{\Pi}_{i}$ and $\boldsymbol{\Pi}_{m}$ to learn the Ego$^2$-Map contrastive objective $\mathcal{L}_{c}$. $f_{s}$ and $f_{t}$ are the paired view features of a trajectory (see \S\ref{subsec:ego2map}). Dash arrows indicate pre-processing without gradient update.}
  \label{fig:navit_arch}
  \vspace{-8pt}
\end{figure*}

\paragraph{Visual Representations for Embodied AI}

Recent years have witnessed a trend of moving away from visual encoders pre-trained for object classification in Embodied AI due to their inefficiency in representing complex real-world scenes or mapping to actions. Instead, large vision-language models such as CLIP which demonstrates strong zero-shot performance across visual domains is widely applied~\cite{cui2022can,radford2021clip}, enhancing the semantic representations in robotic manipulation~\cite{shridhar2022cliport}, assisting 3D trajectory modification and speed control~\cite{bucker2022latte}, benefiting the language-conditioned view-selection problem in VLN~\cite{shah2022lmnav,shen2021benefit} as well as improving other control and navigation results~\cite{gadre2022onwheels,khandelwal2022embclip,parisi2022unsurprising}. Moreover, EmbCLIP shows that the CLIP features provide much better semantic and geometric primitives such as object presence, object reachability, and free space that are valuable to Embodied AI~\cite{khandelwal2022embclip}. In terms of self-supervised visual representation learning for indoor navigation, recent works that are the most relevant to ours include OVRL~\cite{yadav2022ovrl} which applies knowledge distillation based on DINO~\cite{caron2021dino}, EPC which predicts masked observations from a trajectory~\cite{ramakrishnan2021epc}, and CRL which jointly learns a visual encoder with a policy network that maximizes the representation error~\cite{du2021crl}. In contrast to all previous works, this paper explores a novel idea of encoding the explicit structural and semantic information available in maps implicitly in the visual encoder thereby facilitating efficient and effective generalization across different visual navigation tasks and models.

\paragraph{Exploiting Visual Features}

A great variety of proxy tasks and auxiliary objectives have been applied to exploit information from the visual data which is beneficial to navigation. For instance, Ye~\etal~\cite{ye2021auxiliary,ye2021auxpoint} promote understanding of spatiotemporal relations of observations by estimating the timestep difference between two views, Qi~\etal~\cite{qi2021road} predict room-types from object features to improve scene understanding, and Se~\etal~\cite{savinov2018semi} determine the furthest reachable waypoint on a topological graph using the node observations. To extract the geometric information, Gordon~\etal~\cite{gordon2019splitnet} use the encoded visual features to predict depth and surface normals, reconstruct RGB input, and forecast the visual features by taking certain actions. Depth prediction has also been studied by Mirowski~\etal~\cite{mirowski2016complex}, Desai~\etal~\cite{desai2021auxiliary} and Chattopadhyay~\etal~\cite{chattopadhyay2021robustnav}, along with learning the inverse dynamics by predicting an action taken between
two sequential observations~\cite{hansen2020self,mirowski2016complex}. Complementary to the previous methods for interpreting visual features, our Ego$^{2}$-Map learning guides a visual encoder to produce representations with expressive spatial information.

\section{Contrastive Learning between Egocentric Views and Semantic Maps (Ego$^{2}$-Map)}
\label{sec:navit}

We will first describe the overall network architecture for training the visual encoders (\S\ref{subsec:network}), followed by the details of Ego$^{2}$-Map objectives (\S\ref{subsec:ego2map}), as well as two other widely applied spatial-aware auxiliary tasks which we investigated with our Ego$^{2}$-Map learning (\S\ref{subsec:aux_tasks}). Then, we will talk about the data collection (\S\ref{subsec:data_collect}) and network training (\S\ref{subsec:training}) processes.

\subsection{Network Architecture}
\label{subsec:network}

We build the network for Ego$^{2}$-Map learning based on a ViT-B/16 model (default initialized from CLIP~\cite{radford2021clip}, whose effect will be studied in \S{\ref{subsec:discuss}}). Unlike previous methods for visual navigation, which mostly employ two independent encoders to process RGB and depth images, we investigate a compact representation by feeding RGB+Depth as four-channel inputs (Figure~\ref{fig:navit_arch}). To merge the two visual modalities, we use two separate convolutional layers to encode the RGB channels $I_{rgb}$ and the depth map $I_{d}$, respectively, followed by a token-wise concatenation and a non-linear projection $\boldsymbol{\Pi}_{v}$ to merge the resulting feature maps before feeding to the transformer layers as
\begin{equation}
v_{rgb}=\text{Conv}_{rgb}(I_{rgb}),\hspace{20pt}v_{d}=\text{Conv}_{d}(I_{d})
\label{eqn:in_convs}
\end{equation}
and
\begin{equation}
f = \text{ViT}\left(\boldsymbol{\Pi}_{v}\left[v_{rgb};v_{d}\right]\right)
\label{eqn:clip_enc}
\end{equation}
The pooled features $f$ of the RGBD encoder are applied for learning all spatial-aware objectives $\mathcal{L}_{\theta}$, $\mathcal{L}_{d}$ and $\mathcal{L}_{c}$, which will be specified in the following.

\begin{figure*}[t]
  \centering
  \includegraphics[width=\textwidth]{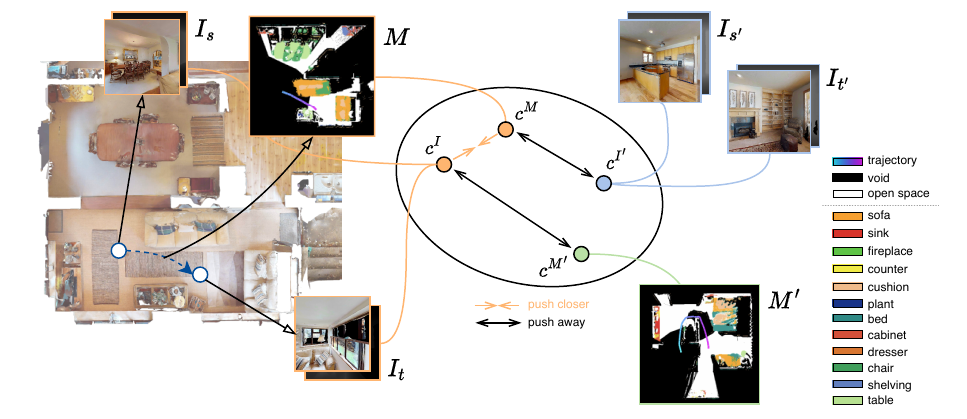}
  \vspace{-20pt}
  \caption{
  Illustration of Ego$^{2}$-Map contrastive learning. 
  The method encodes the paired source and target egocentric views $(I_{s},I_{t})$ into feature $c^{I}$, then learns to align it with the feature $c^{M}$ of the corresponding top-down map $M$. 
  Negative view pairs $(I_{s'},I_{t'})$ and maps ($M'$) are sampled from different trajectories or different environments, which are pushed away from the reference pair. 
  The blue dashed line on the left part of the figure indicates the agent's trajectory, which is visualized with directional color in map $M$. 
  The map is also powered with semantic information (denoted at the bottom right).
  }
  \label{fig:navit_pretrain}
  \vspace{-10pt}
\end{figure*}

\subsection{Ego$^{2}$-Map Contrastive Objective}
\label{subsec:ego2map}

The motivation of this task is to introduce the information within a map to single-view features, offering the agent high-level semantic and spatial clues to navigate towards distant targets. Specifically, we build positive samples by coupling a views-pair ($I_{s}$ and $I_{t}$ from the two endpoints of a path) with a top-down semantic map $M$ that represents the agent's transition and observations along the path, while using mismatched views-pairs and maps of different routes or environments as the negatives. Each semantic map is generated by the off-the-shelf Semantic MapNet~\cite{cartillier2021semmap}, using the RGBD observations collected by an agent traveling from the source view $I_{s}$ to the target view $I_{t}$ via the shortest path\footnote{Each resulting map is unique because the observations are determined by the unique path (actions) for connecting two views. For example, in Figure~\ref{fig:navit_pretrain}, only the colored regions in the map ($M$) are the areas seen by the agent when traveling from the source to the target positions.}. The semantic map provides abundant information about the open space, obstacles, unexplored regions, observed objects, and the agent's action, which we consider as highly valuable and compact representations for agent navigation (see maps in Figure~\ref{fig:navit_pretrain}). To encode $M$, we apply an additional ViT-B/32~\cite{dosovitskiy2020vit} with the last three transformer layers unfrozen to adapt the map images. Then, the egocentric features of the two views $f_{s}$ and $f_{t}$, and the map features $f_{m}$ will be passed to two MLP (multi-layer perceptrons) headers $\boldsymbol{\Pi}_{i}$ and $\boldsymbol{\Pi}_{m}$, respectively, to compute an alignment score as:
\begin{equation}
c^{I}=\boldsymbol{\Pi}_{i}[f_{s};f_{t}],\hspace{20pt}c^{M}=\boldsymbol{\Pi}_{m}[f_{m}]
\label{eqn:contrast_net}
\end{equation}
and
\begin{equation}
\langle{c^{I},c^{M}}\rangle = \frac{c^{I}\hspace{1pt}{\cdot}\hspace{1pt}c^{M}}{\lVert c^{I} \rVert\lVert c^{M} \rVert}
\label{eqn:contrast_score}
\end{equation}
Then, the InfoNCE loss~\cite{oord2018infonce,zhang2020contrastive} is applied for Ego$^{2}$-Map contrastive learning. For each $j$-th views-map pair in a minibatch of size $N$, we have
\begin{equation}
\mathcal{L}_{c,j}=\mathcal{L}^{{I}{\rightarrow}{M}}_{c,j}+\mathcal{L}^{{M}{\rightarrow}{I}}_{c,j}
\label{eqn:info_nce}
\end{equation}
where the views to maps ($I{\rightarrow}M$) loss can be expressed as
\begin{equation}
\mathcal{L}^{{I}{\rightarrow}{M}}_{c,j}=-\text{log}\frac{\text{exp}(\langle{c^{I}_{j},c^{M}_{j}}\rangle/{\tau})}{\sum_{k=1}^{N}\text{exp}(\langle{c^{I}_{j},c^{M}_{k}}\rangle/{\tau})}
\label{eqn:info_nce_i_to_m}
\end{equation}
and the map to views loss $\mathcal{L}^{{M}{\rightarrow}{I}}_{c,j}$ likewise. Parameter $\tau\in\mathbb{R}_{+}$ denotes a learnable temperature.

\subsection{Other Spatial-Aware Objectives}
\label{subsec:aux_tasks}

We investigate two additional widely applied proxy tasks with our Ego$^2$-Map learning. Note that we do not claim any novelty for applying these tasks but focus on studying their effect on training the representations.

\paragraph{Angular Offset Prediction}

Given two images taken from random headings at the same position, the task predicts the angular offset between them to facilitate the modeling of correspondence between views. A similar proxy task has been applied in previous works, which predicts discrete angles and uses an extra directional encoding to imply orientation~\cite{chen2021hamt} or regresses the agent's turning angle conditioned on language during navigation~\cite{zhu2020auxiliary}. Specifically, we pass the two pooled RGBD features ${f}_{\theta_{0}}$ and ${f}_{\theta_{1}}$, corresponding to images at two orientations of the same viewpoint, into a learnable non-linear header $\boldsymbol{\Pi}_{\theta}$ to predict the offset as $\theta^{p}=\boldsymbol{\Pi}_{\theta}[{f}_{\theta_{0}}; {f}_{\theta_{1}}]$. The task is learned by minimizing the mean squared error $\mathcal{L}_{\theta}= E[(\theta^p - \theta^*)^2]$, where $\theta^{*}$ is the ground-truth angular difference between $[-\pi,\pi]$, denoting either clockwise or counter-clockwise rotation whichever is closer to encourage agent's efficient rotation.

\begin{figure}[t]
  \centering
  \includegraphics[width=\columnwidth]{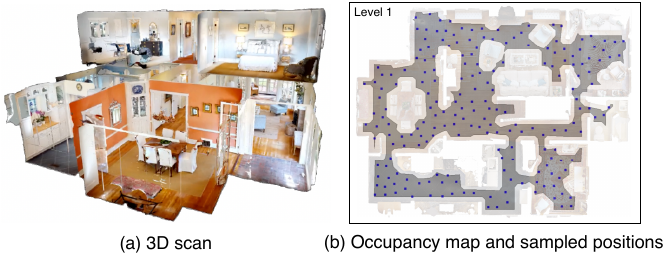}
  \caption{Illustration of an HM3D environment scan and the sampled positions (blue points) in open space (gray area).}
  \label{fig:hm3d_scan}
  \vspace{-10pt}
\end{figure}

\paragraph{Explorable Distance Prediction}

To benefit the searching of explorable regions and assist obstacle avoidance, the task estimates the agent's maximal distance to forward without being blocked by any obstacle~\cite{chaplot2020neural,hahn2021norlnosim}. Using the same notations as above, the module regresses a value from an RGBD image as $d^{p}=\boldsymbol{\Pi}_{d}[{f}]$, and is trained with supervision $\mathcal{L}_{d}=E[(d^{p}-d^{*})^{2}]$. We cap the ground-truth distance $d^{*}$ in the range of 0.5 to 5.0 meters to match the common range of the agent's depth sensors.

\subsection{Data Collection}
\label{subsec:data_collect}

We follow the train-validation-test split of the HM3D scenes~\cite{ramakrishnan2021hm3d} and use the first 800 environments for learning. We randomly sample 252,537 viewpoint positions from the environments (see Figure~\ref{fig:hm3d_scan}) and render more than a million RGBD images from those positions to create 500,000 $(I_{\theta_{0}},I_{\theta_{1}})$ views pairs as well as 500,000 $(I_{s},I_{t},M)$ triplets for learning $\mathcal{L}_{\theta}$ and $\mathcal{L}_{c}$, respectively, while using all the images to learn $\mathcal{L}_{d}$. The positions are sampled such that all points must be located in the open space and the minimal geodesic distance between any two points is greater than 0.40 meters. Each image collected for Ego$^{2}$-Map contrastive learning is either a unique source or target view in all trajectories, and each trajectory is created by computing the shortest path between two randomly paired views within a 7 meters range. We feed the RGBD images captured from a trajectory to the Semantic MapNet~\cite{cartillier2021semmap} for drawing the corresponding semantic map. Note that the Semantic MapNet is trained on the MP3D environments~\cite{chang2017matterport3d}, but we found that it also generalizes well to the HM3D scenes. We refer to the \textit{Appendix} for more sampling and dataset creation details.

\subsection{Training}
\label{subsec:training}

\paragraph{Implementation Details}

We initialize the RGBD encoder and the map encoder of our visual representation network with CLIP, a visual transformer pre-trained for aligning image-language pairs~\cite{radford2021clip}. Specifically, the additional depth convolutional layer and the projection layer in the RGBD encoder are initialized randomly. All network variants in our experiments are trained with batch size 994 for 100 epochs using the AdamW optimizer~\cite{loshchilov2018adamw}. After all training iterations, only the RGBD encoder will be applied as the visual backbone for navigation agents. Image augmentations are applied to all views and maps. 

\paragraph{Optimization}

The overall pre-training strategy minimizes the sum over all three losses, $\mathcal{L}_{c}$, $\mathcal{L}_{\theta}$ and $\mathcal{L}_{d}$.
All losses are equally weighted and optimized simultaneously in each iteration. After all training steps, evaluate the model with batch size 128 on 10,000 novel 
$(I_{s},I_{t},M)$ triplets shows 92.02\% views to maps ($I{\rightarrow}M$) and 92.03\% maps to view ($M{\rightarrow}I$) alignment accuracy, indicating the correspondence between two modalities has been learned. Please see \textit{Appendix} for more pre-training statistics and results.

\section{Downstream Experiments}
\label{sec:experiments}

We evaluate our Ego$^2$-Map representations mainly on the R2R-CE navigation~\cite{anderson2018vln,krantz2020vlnce}, as addressing the task highly relies on exploiting the semantic and structural information from observations and ground to language instructions. We also show results on ObjNav~\cite{batra2020objectnav} to compare with other recent visual representation learning approaches.

\begin{table*}[t]
  \begin{center}
  \resizebox{\textwidth}{!}{
  \begin{tabular}{l|cc|c|rrrr|rrrrr}
    \hline \hline
    \multicolumn{1}{c|}{\multirow{2}{*}{Models}} & \multicolumn{3}{c}{Pre-Training Objectives} & \multicolumn{4}{|c}{R2R-CE Val-Unseen ($\mathcal{A}_{High}$)} & \multicolumn{5}{|c}{R2R-CE Val-Unseen ($\mathcal{A}_{Low}$)} \Tstrut\\
    \cline{2-13} &
    \multicolumn{1}{c}{Angular} & \multicolumn{1}{c|}{Explorable} & \multicolumn{1}{c}{Contrastive} & \multicolumn{1}{|c}{NE$\downarrow$} & \multicolumn{1}{c}{nDTW$\uparrow$} &
    \multicolumn{1}{c}{SR$\uparrow$} & \multicolumn{1}{c}{SPL$\uparrow$} & \multicolumn{1}{|c}{NE$\downarrow$} & \multicolumn{1}{c}{nDTW$\uparrow$} &
    \multicolumn{1}{c}{SR$\uparrow$} & \multicolumn{1}{c}{SPL$\uparrow$} & \multicolumn{1}{c}{Collision$\downarrow$} \Tstrut\\
    \hline \hline
    Baseline &   &   &   & 4.98 & 57.26 & 48.67 & 42.55 & 7.69 & 49.27 & 25.61 & 23.93 & 17.41 \\
    \hline
    1 & \checkmark &  &  & 8.33 & 32.02 & 16.15 & 13.65 & 9.77 & 30.43 & 3.75 & 3.43 & 71.59 \\
    2 &  & \checkmark &  & 5.83 & 51.25 & 42.58 & 36.34 & 7.94 & 47.15 & 24.74 & 22.98 & 10.44 \\
    3 &  &  & \checkmark & 4.89 & 58.49 & 51.77 & 45.89 & 7.47 & 50.35 & 27.62 & 25.95 & 13.19 \\
    \hline
    4 &  & \checkmark & \checkmark & 5.03 & 58.86 & 51.28 & 45.98 & 7.27 & 51.91 & 30.61 & 29.13 & 10.09 \\
    5 & \checkmark &  & \checkmark & 5.02 & 59.62 & 52.04 & 46.99 & 7.44 & 51.31 & 29.53 & 27.94 & 13.04 \\
    6 & \checkmark & \checkmark &  & 5.79 & 52.93 & 42.90 & 37.25 & 7.64 & 50.19 & 27.95 & 26.39 & 16.19 \\
    \hline
    7 & \checkmark & \checkmark & \checkmark & 4.94 & 59.67 & 51.77 & 46.11 & 7.25 & 52.01 & 30.40 & 29.03 & 11.25 \\
    \hline \hline
  \end{tabular}}
\end{center}
\vspace{-10pt}
\caption{Ablation of Ego$^2$-Map learning and influence of angular and explorable objectives on the R2R-CE tasks. Checkmarks indicate applying the corresponding objectives. \textit{Baseline} applies a pre-trained CLIP-ViT-B/16 model to encode RGB and depth separately. \textit{Collision} measures the percentage of forward steps which the agent collided with an obstacle; we only consider this metric for $\mathcal{A}_{Low}$ since agents in $\mathcal{A}_{High}$ apply a waypoint predictor~\cite{hong2022bridging} to obtain navigable positions.}
\label{tab:abl_obj}
\vspace{-10pt}
\end{table*}

\subsection{R2R-CE Setup}


Our experiments consider two major setups of action spaces, either applying panorama inputs and execute high-level decisions by selecting images pointing towards navigable directions ($\mathcal{A}_{High}$)\footnote{High-level VLN agents apply 36 egocentric images to represent its panoramic vision~\cite{anderson2018vln}, hence, our Ego$^2$-Map features are applicable.}, or using egocentric views and perform low-level controls ($\mathcal{A}_{Low}$). For $\mathcal{A}_{High}$, we adopt the candidate waypoint predictor proposed by Hong~\etal~\cite{hong2022bridging}, which generates navigable waypoint positions around the agent at each time step. Once a waypoint is selected, the agent will move to the position with low-level controls. For $\mathcal{A}_{Low}$, an agent is only allowed to do a single low-level action at each step, including Turn Left/Right 15$^{\circ}$, Forward 0.25 meters, and Stop. In all experiments, we apply the \vlnbert~agent~\cite{hong2020recbert}, a vision-language transformer pre-trained for navigation, as the policy network. To use our Ego$^2$-Map representations, we simply replace the agent's original visual encoders (two ResNets trained on ImageNet~\cite{he2016resnet} and trained in Point-Nav~\cite{wijmans2019ddppo} for RGB and depth, respectively) with our RGBD encoder (\S\ref{subsec:network}). Comparing the runtime efficiency, the original encoder has around 50M parameters and the speed is about 4 GFLOPs/image, whereas our method is larger (86M parameters) but faster (20 GFLOPs/image).
All visual encoders in our experiments are frozen in navigation.

R2R in continuous environment (R2R-CE) is established over the Matterport3D environments~\cite{chang2017matterport3d} based on the Habitat simulator~\cite{savva2019habitat}. The dataset contains 61 scans for training; 11 and 18 scans for validation and testing, respectively. R2R-CE evaluates the agent's performance using multiple metrics, including trajectory length (TL), navigation error (NE): the final distance between the agent and the target, normalized dynamic time warping score (nDTW): distance between the predicted and the ground-truth paths~\cite{ilharco2019general}, oracle success rate (OSR): the ratio of reaching within 3 meters to the target, success rate (SR): the ratio of stopping within 3 meters to the target, and success weighted by the normalized inverse of the path length (SPL)~\cite{anderson2018evaluation}.

\subsection{Ablation Study}

Ablation experiments in Table~\ref{tab:abl_obj} reveal the influence of different pre-training objectives. We establish our Baseline by using a frozen CLIP-ViT-B/16 to encode RGB and depth separately, followed by a trainable fusion layer to merge the two features, which is a strong baseline that can achieve better results than the previous best encoders (see CLIP-ViT+Depth in Table~\ref{tab:q_model0}).

Results of Model\#1 indicate that employing angular offset prediction as the only task has a devastating effect on the encoder; in fact, we found that $\mathcal{L}_{\theta}$ only oscillates if it is minimized alone, leading to features that are not useful in downstream tasks (see details in \textit{Appendix}). Although learning $\mathcal{L}_{\theta}$ with $\mathcal{L}_{d}$ or $\mathcal{L}_{c}$ can improve the performance in both $\mathcal{A}_{High}$ and $\mathcal{A}_{Low}$ (Model\#5, Model\#6), removing the task from Model\#7 will not cause a noticeable difference (Model\#4). 
Meanwhile, learning explorable distance prediction (Model\#2, Model\#5) is not effective in $\mathcal{A}_{High}$ scenario because agents with $\mathcal{A}_{High}$ apply pre-defined waypoints on open space, which means, finding explorable directions and avoid obstacle are not necessary. However, Model\#2, Model\#4 and Model\#6 suggest that learning $\mathcal{L}_{d}$ with $\mathcal{L}_{c}$ or $\mathcal{L}_{\theta}$ will boost the results in $\mathcal{A}_{Low}$ and effectively reduce the collision rate during navigation.

On the other hand, by comparing the Baseline, Model\#3, Model\#6, and Model\#7, we can see that our proposed Ego$^{2}$-Map contrastive learning has the largest impact on R2R-CE. Solely learning $\mathcal{L}_{c}$ improves the SR and SPL absolutely by 3.10\% and 3.34\% in $\mathcal{A}_{High}$, as well as 2.01\% and 2.02\% in $\mathcal{A}_{Low}$, while removing $\mathcal{L}_{c}$ from Model\#7 dramatically lowers the results in both settings. Meanwhile, we found that Ego$^{2}$-Map learning can also help avoid collision, implying the useful spatial information carried by the map. In the following experiments, all three losses are applied to pre-train the visual encoders.

\subsection{Discussion and Analysis}
\label{subsec:discuss}

\paragraph{How much does the quantity of visual data matter?}

\begin{table}[t]
  \begin{center}
  \resizebox{\columnwidth}{!}{
  \begin{tabular}{l|rrrr|rrrr}
    \hline \hline
    \multicolumn{1}{c|}{\multirow{2}{*}{Pre-train Data}} & \multicolumn{4}{c}{R2R-CE Val-Unseen ($\mathcal{A}_{High}$)} & \multicolumn{4}{|c}{R2R-CE Val-Unseen ($\mathcal{A}_{Low}$)} \\
    \cline{2-9} & 
    \multicolumn{1}{c}{NE$\downarrow$} & \multicolumn{1}{c}{nDTW$\uparrow$} &
    \multicolumn{1}{c}{SR$\uparrow$} & \multicolumn{1}{c}{SPL$\uparrow$} & \multicolumn{1}{|c}{NE$\downarrow$} & \multicolumn{1}{c}{nDTW$\uparrow$} &
    \multicolumn{1}{c}{SR$\uparrow$} & \multicolumn{1}{c}{SPL$\uparrow$} \Tstrut\\
    \hline \hline
    100\% data & 4.94 & 59.67 & 51.77 & 46.11 & 7.25 & 52.01 & 30.40 & 29.03 \\
    \hline
    50\% samples & 4.98 & 57.72 & 50.46 & 45.03 & 7.43 & 49.97 & 27.57 & 26.22 \\
    30\% samples & 5.28 & 56.11 & 48.61 & 42.58 & 7.83 & 48.60 & 25.83 & 24.63 \\
    10\% samples & 5.60 & 54.89 & 44.48 & 39.08 & 7.86 & 46.83 & 22.95 & 21.54 \\
    \hline
    50\% envs & 5.32 & 55.53 & 47.80 & 42.24 & 7.57 & 50.35 & 26.54 & 25.15 \\
    30\% envs & 5.56 & 53.81 & 44.48 & 38.69 & 7.61 & 48.64 & 26.05 & 24.51 \\
    10\% envs & 5.73 & 53.44 & 41.92 & 36.89 & 7.88 & 47.62 & 23.22 & 22.00 \\
    \hline
    MP3D & 5.67 & 52.78 & 40.78 & 35.76 & 7.97 & 46.75 & 21.97 & 20.60 \\
    \hline \hline
  \end{tabular}}
\end{center}
\vspace{-10pt}
\caption{Effect of additional environments for pre-training. The percentages indicate the portion of HM3D environments/samples for training. MP3D only uses data collected from the 61 environments in downstream.}
\label{tab:q_data}
\vspace{-5pt}
\end{table}

In Table~\ref{tab:q_data}, we compare the effect of sampling a different quantity of data for pre-training our visual encoder. Results show that the agent's performance in R2R-CE drops as the amount of data decreases, and reducing the number of environments has a stronger impact on the results (-1.08\% SPL for 50\% samples vs. -3.87\% SPL for 50\% envs with $\mathcal{A}_{High}$), suggesting the importance of having abundant scenes and structures in learning visual representations.
Meanwhile, we can see that applying less than 50\% of samples or environments cannot yield better results than the baseline in Table~\ref{tab:abl_obj}. This is because Ego$^2$-Map learning requires sufficient positive and negative data pairs to support its contrastive training, as observed in previous literatures~\cite{chen2020simclr,radford2021clip}.

\paragraph{How much does initialization matter?}

\begin{table}[t]
  \begin{center}
  \resizebox{\columnwidth}{!}{
  \begin{tabular}{l|rrrr|rrrr}
    \hline \hline
    \multicolumn{1}{c|}{\multirow{2}{*}{Initialization}} & \multicolumn{4}{c}{R2R-CE Val-Unseen ($\mathcal{A}_{High}$)} & \multicolumn{4}{|c}{R2R-CE Val-Unseen ($\mathcal{A}_{Low}$)} \\
    \cline{2-9} & 
    \multicolumn{1}{c}{NE$\downarrow$} & \multicolumn{1}{c}{nDTW$\uparrow$} &
    \multicolumn{1}{c}{SR$\uparrow$} & \multicolumn{1}{c}{SPL$\uparrow$} & \multicolumn{1}{|c}{NE$\downarrow$} & \multicolumn{1}{c}{nDTW$\uparrow$} &
    \multicolumn{1}{c}{SR$\uparrow$} & \multicolumn{1}{c}{SPL$\uparrow$} \Tstrut\\
    \hline \hline
    CLIP & 4.94 & 59.67 & 51.77 & 46.11 & 7.25 & 52.01 & 30.40 & 29.03 \\
    \hline
    Random & 4.99 & 57.55 & 49.32 & 43.89 & 7.27 & 51.35 & 29.91 & 28.29 \\
    \hline
    IN-21K & 4.95 & 58.61 & 51.55 & 45.73 & 7.05 & 53.30 & 31.05 & 29.69 \\
    +IN-21K map & 5.40 & 56.18 & 47.09 & 42.01 & 7.21 & 52.68 & 28.55 & 27.13 \\
    \hline \hline
  \end{tabular}}
\end{center}
\vspace{-10pt}
\caption{Effect of using pre-trained encoders for RGBD and map. \textit{Random} and \textit{IN-21K} initialize the RGBD encoder randomly or from a ViT pre-trained on ImageNet-21K~\cite{deng2009imagenet}, while using the CLIP-ViT-B/32 to encode maps. \textit{+IN-21K map} also uses the IN-21K pre-trained ViT to encode the semantic map.}
\label{tab:q_init_clip}
\vspace{-15pt}
\end{table}

Table~\ref{tab:q_init_clip} compares the initialization of the RGBD and the map encoders in Ego$^2$-Map network, either from CLIP~\cite{radford2021clip}, from scratch or from a ViT-B/16~\cite{dosovitskiy2020vit} pre-trained on ImageNet-21K (IN-21K)~\cite{deng2009imagenet}. We find that random initialization of the RGB encoder only leads to a slight decrease in R2R-CE performance, while initializing it from pre-trained ViT can achieve comparable results to Model\#7. We further investigate these results by also replacing the initialization of our map encoder from CLIP with IN-21K pre-trained ViT; although the losses of the two models saturate to the same level in pre-training, a drastic drop is observed in R2R-CE (even lower than \textit{Random}). We hypothesize that some semantics from CLIP can be introduced to the RGBD encoder by the map encoder through contrastive learning. While previous works often use a convolutional network to process maps~\cite{chen2022weakly,georgakis2021learning,georgakis2022cross,ramakrishnan2022poni}, there remains a valuable open question of what model is suitable for encoding maps and how to train it to benefit navigation. We will leave the study of encoding maps for future work.

\paragraph{Effect of $\mathcal{L}$ on a different encoder}

\begin{table}[t]
  \begin{center}
  \resizebox{\columnwidth}{!}{
  \begin{tabular}{l|rrrr|rrrr}
    \hline \hline
    \multicolumn{1}{c|}{\multirow{2}{*}{Methods}} & \multicolumn{4}{c}{R2R-CE Val-Unseen ($\mathcal{A}_{High}$)} & \multicolumn{4}{|c}{R2R-CE Val-Unseen ($\mathcal{A}_{Low}$)} \\
    \cline{2-9} & 
    \multicolumn{1}{c}{NE$\downarrow$} & \multicolumn{1}{c}{nDTW$\uparrow$} &
    \multicolumn{1}{c}{SR$\uparrow$} & \multicolumn{1}{c}{SPL$\uparrow$} & \multicolumn{1}{|c}{NE$\downarrow$} & \multicolumn{1}{c}{nDTW$\uparrow$} &
    \multicolumn{1}{c}{SR$\uparrow$} & \multicolumn{1}{c}{SPL$\uparrow$} \Tstrut\\
    \hline \hline
    CLIP-ViT+Depth & 5.47 & 55.55 & 47.15 & 41.28 & 7.93 & 47.76 & 22.68 & 21.69 \\    
    +$\mathcal{L}$ & 5.24 & 57.74 & 48.67 & 43.78 & 7.07 & 52.59 & 30.07 & 28.74 \\
    \hline \hline
  \end{tabular}}
\end{center}
\vspace{-10pt}
\caption{Pre-training a different visual encoder with Ego$^2$-Map objectives. The model applies CLIP-ViT-B/16
and a ResNet-50 trained for Point-Nav~\cite{wijmans2019ddppo} to encode RGB and depth, respectively.}
\label{tab:q_model0}
\vspace{-10pt}
\end{table}

We further evaluate our proposed Ego$^2$-Map objective in pre-training a different visual encoder (Table~\ref{tab:q_model0}). Here we consider the previous state-of-the-art encoder CLIP-ViT+Depth, which applies CLIP-ViT-B/16 and a ResNet-50 depth net trained on Gibson~\cite{xia2018gibson} for PointNav~\cite{savva2019habitat} to encode RGB and depth, respectively, following by a fusion module to integrate the encoded features. Pre-training the model with $\mathcal{L}$ leads to a significant improvement, obtaining +2.50\% and +7.05\% higher SPL with $\mathcal{A}_{High}$ and $\mathcal{A}_{Low}$. These results suggest that Ego$^2$-Map learning has the potential to be generalized to other visual encoders.

\paragraph{Information in Ego$^{2}$-Map contrastive learning}

In Table~\ref{tab:q_map_info}, we ablate the information applied in the Ego$^{2}$-Map contrastive learning. Specifically, we either remove the object semantics by masking their colored segmentations with the same color, or remove explorable regions by masking open space and void with the same color (see unmasked maps in Figure~\ref{fig:navit_pretrain}), or remove the target view from the view pairs to create ambiguity in agent's transition. The results show that learning Ego$^{2}$-Map without the semantic clues or open space in the maps, or without a specified target greatly damages agent performance in $\mathcal{A}_{High}$, reflecting the importance of including this information. We also find that the agent with $\mathcal{A}_{Low}$ is less sensitive to method variations in pre-training; the remaining spatial or object information on a map is still beneficial to enhance the visual representation for supporting $\mathcal{A}_{Low}$ navigation, which could be the reason why angular offset and explorable distance predictions do not show a consistent benefit in Table~\ref{tab:abl_obj}. 

\begin{table}[t]
  \begin{center}
  \resizebox{\columnwidth}{!}{
\begin{tabular}{l|rrrr|rrrr}
    \hline \hline
    \multicolumn{1}{c|}{\multirow{2}{*}{Ego$^{2}$-Map}} & \multicolumn{4}{c}{R2R-CE Val-Unseen ($\mathcal{A}_{High}$)} & \multicolumn{4}{|c}{R2R-CE Val-Unseen ($\mathcal{A}_{Low}$)} \\
    \cline{2-9} & 
    \multicolumn{1}{c}{NE$\downarrow$} & \multicolumn{1}{c}{nDTW$\uparrow$} &
    \multicolumn{1}{c}{SR$\uparrow$} & \multicolumn{1}{c}{SPL$\uparrow$} & \multicolumn{1}{|c}{NE$\downarrow$} & \multicolumn{1}{c}{nDTW$\uparrow$} &
    \multicolumn{1}{c}{SR$\uparrow$} & \multicolumn{1}{c}{SPL$\uparrow$} \Tstrut\\
    \hline \hline
    No semantics & 5.51 & 57.60 & 47.04 & 42.25 & 7.31 & 51.56 & 29.74 & 28.34 \\
    No space & 5.24 & 57.42 & 48.94 & 43.52 & 7.66 & 49.20 & 29.31 & 27.75 \\    
    No target & 5.02 & 56.26 & 47.20 & 41.21 & 8.05 & 51.04 & 30.83 & 29.07 \\
    \hline \hline
  \end{tabular}}
\end{center}
\vspace{-10pt}
\caption{Ablation of information in Ego$^{2}$-Map learning.}
\label{tab:q_map_info}
\vspace{-10pt}
\end{table}

\subsection{Comparison to Previous Methods}

\paragraph{Advantages of Visual Representations (R2R-CE)}
In Table~\ref{tab:main_results_r2r}, we compare our method of applying Ego$^2$-Map RGBD encoder to the CWP-\vlnbert~agent~\cite{hong2022bridging} (same model as Model\#7 in Table~\ref{tab:abl_obj}) with previous approaches using $\mathcal{A}_{High}$ on the R2R-CE testing split\footnote{R2R-CE Challenge Leaderboard: \href{https://eval.ai/web/challenges/challenge-page/719/leaderboard/1966}{https://eval.ai/web/challenges/challenge-page/719/leaderboard/1966}}. Results show that our proposed Ego$^2$-Map learning brings significant improvement for all metrics, achieving a 47\% SR (+3\%) and a 41\% SPL (+4\%) over the previous best~\cite{krantz2022sim}. In addition, Table~\ref{tab:compare_map} establishes a fair comparison with methods using $\mathcal{A}_{Low}$. We can see that Ego$^2$-Map features greatly boost the result of the base agent CWP-\vlnbert~\cite{hong2022bridging} (23\% to 30\% SR), and achieve a SPL comparable to recent mapping-based methods such as CM2~\cite{georgakis2022cross} and WS-MGMap~\cite{chen2022weakly}. Note that using Ego$^2$-Map representations does not conflict with online mapping; while we only experiment with non-mapping-based agents, we believe the features hold great potential to facilitate modeling between views and maps in mapping-based models.

\paragraph{Learning Methods (ObjNav)}
We further compare the effect of visual representation learning methods by applying our visual encoder on ObjNav~\cite{batra2020objectnav} as in the recent approaches~\cite{khandelwal2022embclip,yadav2022ovrl} (Table~\ref{tab:main_results_objnav}). All methods in the table adopt a simple pipeline as in the baseline model~\cite{wijmans2019ddppo}, which feeds the concatenation of visual features, GPS+Compass encodings, and the encodings of previous action to a GRU~\cite{cho2014gru}, then, uses a fully-connected layer to predict an action from the updated agent's state. Briefly, EmbCLIP directly uses the pre-trained CLIP-ResNet50~\cite{radford2021clip} to encode the RGB inputs, EmbCLIP-ViT+Depth applies the CLIP-ViT-B/16 to encode RGB and an extra depth net~\cite{savva2019habitat} pre-trained on Gibson~\cite{xia2018gibson} to encode the depth inputs. OVRL pre-trains the ResNet encoder with self-distillation method DINO~\cite{caron2021dino}, in which a student network is trained to match the output of a teacher network. Moreover, OVRL applies the Omnidata Starter Dataset (OSD)\footnote{OSD~\cite{eftekhar2021omnidata} contains approximately 14.5 million images rendered from diverse 3D environments, including Replica~\cite{straub2019replica}, Replica+GSO~\cite{downs2022gso}, Hypersim~\cite{roberts2021hypersim}, Taskonomy~\cite{zamir2018taskonomy}, BlendedMVG~\cite{yao2020blendedmvs} and HM3D~\cite{ramakrishnan2021hm3d}.}~\cite{eftekhar2021omnidata}, which is much larger and more diverse than the data for our Ego$^2$-Map learning (only uses the HM3D~\cite{ramakrishnan2021hm3d} subset). ObjNav measures the same metrics as R2R-CE, first, we can see that an agent using Ego$^2$-Map features greatly improves over the baseline and the EmbCLIP. Compared to OVRL, despite the method applies OSD for pre-training and fine-tunes the network end-to-end with human demonstrations, our method obtains a better success rate (+0.4\%) and much higher SPL (+3.2\%). Note that, although not directly comparable, as reported in the OVRL, the SPL on ImageNav~\cite{mezghani2021imagenav} without fine-tuning the visual encoder will drop drastically from 26.9\% to 17.0\%, whereas all our experiments keep the visual encoder frozen during navigation. These results suggest that Ego$^2$-Map representations are generalizable to different navigation tasks and provide more robust visual representations.

\begin{table}[t!]
  \begin{center}
  \resizebox{\columnwidth}{!}{
  \begin{tabular}{l|ccccc}
    \hline \hline
    \multicolumn{1}{c|}{\multirow{2}{*}{Methods ($\mathcal{A}_{High}$)}} & 
    \multicolumn{5}{c}{R2R-CE Test-Unseen} \\
    \cline{2-6} &  \multicolumn{1}{c}{TL} & \multicolumn{1}{c}{NE$\downarrow$} &
    \multicolumn{1}{c}{OSR$\uparrow$} & \multicolumn{1}{c}{SR$\uparrow$} & \multicolumn{1}{c}{SPL$\uparrow$} \Tstrut\\
    \hline \hline
    Waypoint Models~\cite{krantz2021waypoint} & 8.02 & 6.65 & 37 & 32 & 30 \\
    CWP-CMA~\cite{hong2022bridging} & 11.85 & 6.30 & {49} & {38} & {33} \\
    CWP-\vlnbert~\cite{hong2022bridging} & 13.31 & 5.89 & 51 & 42 & 36 \\
    Sim2Sim~\cite{krantz2022sim} & 11.43 & 6.17 & 52 & 44 & 37 \\
    \hline
    Ego$^2$-Map+CWP-\vlnbert~(ours) & 13.05 & \textbf{5.54} & \textbf{56} & \textbf{47} & \textbf{41} \\
    \hline \hline
  \end{tabular}}
\end{center}
\vspace{-15pt}
\caption{Comparison of agent performance on R2R-CE test server. All methods use high-level action space ($\mathcal{A}_{High}$).}
\label{tab:main_results_r2r}
\vspace{-10pt}
\end{table}

\begin{table}[t!]
  \begin{center}
  \resizebox{\columnwidth}{!}{
  \begin{tabular}{l|ccccc}
    \hline \hline
    \multicolumn{1}{c|}{\multirow{2}{*}{Methods ($\mathcal{A}_{Low}$)}} & 
    \multicolumn{5}{c}{R2R-CE Val-Unseen} \\
    \cline{2-6} &  \multicolumn{1}{c}{TL} & \multicolumn{1}{c}{NE$\downarrow$} &
    \multicolumn{1}{c}{OSR$\uparrow$} & \multicolumn{1}{c}{SR$\uparrow$} & \multicolumn{1}{c}{SPL$\uparrow$} \Tstrut\\
    \hline \hline
    CMA+PM+DA+Aug~\cite{krantz2020vlnce} & 8.64 & 7.37 & 40 & 32 & 30 \\
    SASRA~\cite{irshad2021sasra} $\dag$ & 7.89 & 8.32 & -- & 24 & 22 \\
    LAW~\cite{raychaudhuri2021law} & 8.89 & 6.83 & \textbf{44} & \textbf{35} & \textbf{31} \\
    CWP-CMA~\cite{hong2022bridging} $^{\circ}$ & 8.22 & 7.54 & -- & 27 & 25 \\
    CWP-\vlnbert~\cite{hong2022bridging} $^{\circ}$ & 7.42 & 7.66 & --  & 23 & 22 \\
    CM2~\cite{georgakis2022cross} $\dag$ & 11.54 & 7.02 & 42 & 34 & 28 \\
    WS-MGMap (SemMap only)~\cite{chen2022weakly} $\dag$ & 10.89 & \textbf{6.80} & 42 & 33 & 28 \\
    \hline
    Ego$^2$-Map+CWP-\vlnbert~(ours) & 8.03 & 7.25 & 37 & 30 & 29 \\
    \hline \hline
  \end{tabular}}
\end{center}
\vspace{-10pt}
\caption{Comparison of agents with low-level action space ($\mathcal{A}_{Low}$) in R2R-CE Val-Unseen. $\dag$: mapping-based methods. $^{\circ}$: 4\% of data is removed but the comparison is still valid due to the large performance gap.}
\label{tab:compare_map}
\vspace{-5pt}
\end{table}

\begin{table}[t!]
  \begin{center}
  \resizebox{\columnwidth}{!}{
  \begin{tabular}{l|c|rrr}
    \hline \hline
    \multicolumn{1}{c|}{\multirow{2}{*}{Methods}} & \multicolumn{1}{c|}{\multirow{2}{*}{Pre-Training Dataset}} & \multicolumn{3}{c}{ObjNav MP3D Val} \\
    \cline{3-5} & & \multicolumn{1}{c}{NE$\downarrow$} & \multicolumn{1}{c}{SR$\uparrow$} &
    \multicolumn{1}{c}{SPL$\uparrow$} \Tstrut\\
    \hline \hline
    Baseline~\cite{wijmans2019ddppo} & \xmark & 6.90 & 8.0 & 1.8 \\
    EmbCLIP~\cite{khandelwal2022embclip}$^{*}$ & WebImageText & 5.26 & 20.9 & 8.3 \\
    EmbCLIP-ViT+Depth & WebImageText+Gibson & \textbf{4.90} & 23.3 & 8.6 \\
    OVRL no pretrain~\cite{ramrakhya2022habweb} $\dag$ & \xmark & -- & 24.2 & 5.9 \\
    OVRL~\cite{yadav2022ovrl} $\dag$ & OSD & -- & 28.6 & 7.4 \\
    \hline
    Ego$^2$-Map+Baseline (ours) & HM3D & 5.17 & \textbf{29.0} & \textbf{10.6} \\
    \hline \hline
  \end{tabular}}
\end{center}
\vspace{-10pt}
\caption{Comparison on pre-trained visual encoders for ObjNav. $\dag$ indicates the visual encoder is tuned end-to-end with behavior cloning on 40k human demonstrations collected by Habitat-Web~\cite{ramrakhya2022habweb}, while the others freeze the visual encoder during navigation and only train the agent with DD-PPO~\cite{wijmans2019ddppo} on the original data. $^{*}$Results obtained by re-evaluating the officially released best model checkpoint.}
\label{tab:main_results_objnav}
\vspace{-10pt}
\end{table}

\section{Conclusion}
\label{sec:conclusion}

In this paper, we introduce a novel method of learning navigational visual representations with contrastive learning between egocentric views pairs and top-down semantic maps (Ego$^{2}$-Map). The method transfers the compact semantic and spatial information carried by a map to the egocentric representations, which greatly facilities the agent's visual perception. Experiments show that Ego$^{2}$-Map features greatly improve the downstream navigation, such as ObjNav and VLN, and demonstrate generalization potential to different visual backbones. We believe the Ego$^{2}$-Map contrastive learning proposes a new direction of visual representation learning for navigation and provides the possibility of better modeling the correspondence between views and maps, which can further benefit agent's planning and action. Note that our work also produces a potentially effective map encoder whose full capability is worth investigating in future work. 

\vspace{-10pt}

\paragraph{Limitations}
The need to build semantic maps to enable the Ego$^{2}$-Map contrastive learning is an inevitable cost of this method; compared to other self-supervised visual representation learning approaches, Ego$^{2}$-Map requires either the semantic annotations of scenes, or traversable environments and a generalizable semantic map constructor. As a result, the data collection could be much harder.
However, we also witnessed an increasing number of interactive 3D scenes being built in recent years with dense semantic annotations~\cite{deitke2022procthor,ramakrishnan2021hm3d,savva2019habitat,straub2019replica,xia2018gibson}, which can facilitate scaling up our Ego$^{2}$-Map learning in the future.



{\small
\bibliographystyle{ieee_fullname}
\bibliography{egbib}
}

\clearpage

\appendix
\section*{Appendices}

\section{Data Sampling}

We sample images and create top-down semantic maps for our Ego$^2$-Map pre-training from the large-scale Habitat-Matterport3D environments (HM3D)~\cite{ramakrishnan2021hm3d,savva2019habitat}. HM3D provides 1,000 high-fidelity reconstructions of entire buildings, containing indoor spaces at diverse geographical locations and physical sizes, with a total traversable area of 112.5$k$ m$^{2}$. HM3D environments are divided into 800 training, 100 validation, and 100 testing scenes. In our work, we sample data from the training and validation scenes for pre-training and evaluation, respectively, and the sampling details will be provided in this section.

\subsection{Viewpoint Sampling}

We sample viewpoints from the environments using a virtual agent in the Habitat simulator~\cite{savva2019habitat}. We initialize the agent such that its physical dimensions match the standard configurations in R2R-CE~\cite{krantz2020vlnce} (0.10 m radius and 1.50 m height with a camera pointing horizontally at 1.25 m height). The setting of physical dimensions determines the virtual agent's navigable space, which is also the open space that we applied to sample the viewpoints.

We create a heuristic to control the sampling process; First, for each scene, we measure the size of the total navigable area $\mathcal{S}$ using the \texttt{sim.pathfinder.navigable\_area()} function\footnote{All functions correspond to the functions defined in the Habitat Simulator, please refer to their detailed definitions in the official codebase: \href{https://github.com/facebookresearch/habitat-sim}{https://github.com/facebookresearch/habitat-sim}.}. Then, we apply \texttt{sim.sample\_navigable\_point()} to randomly sample a viewpoint $p$ in the open space. Each $p$ needs to satisfy three criteria otherwise discarded, checking by three functions: (1) \texttt{sim.sample\_navigable\_point($p$)}: its position should be reachable from all other open regions, (2) \texttt{sim.island\_radius($p$)}$<$1.50: the point should not be sampled in narrow spaces, and (3) \texttt{sim.geodesic\_distance($p_i$, $p_j$)}$<$0.40: the minimal distance between any two viewpoints should be greater than 0.40 meters. For each scene, we repeat the above procedure to sample 4$\times\mathcal{S}$ or 500 viewpoints\footnote{A maximum is necessary because HM3D contains a few giant scans with repetitive spaces (\textit{e.g.} hotel rooms), which only offer very little image diversity so that are inappropriate for training.}, whichever is smaller.

\begin{figure*}[t]
  \centering
  \includegraphics[width=\textwidth]{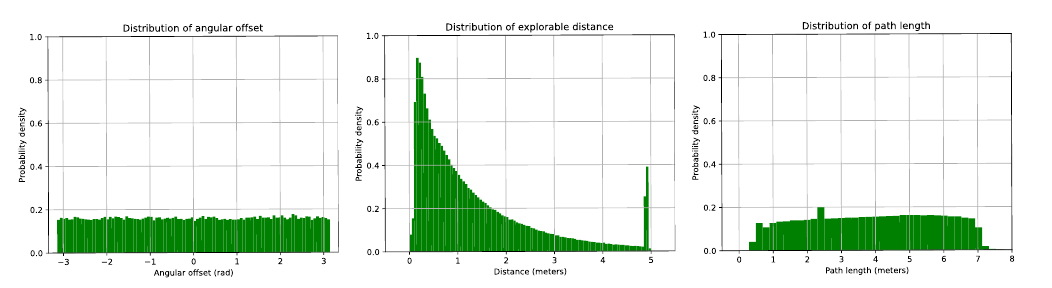}
  \caption{Distribution of the sampled data. Left: angular offset for views pairs. Middle: explorable distance. Right: path length between source and target viewpoints (computed by the \texttt{ShortestPathFollower()} function).
  }
  \label{fig:data_stats}
\end{figure*}

\subsection{Image Sampling}

For each valid viewpoint, we use the camera to capture four RGBD images with a resolution of 320$\times$320 and a horizontal field of view of 90$^{\circ}$, where the range of depth sensor is $[0.5,5.0]$ meters. The views are sampled at random orientations, and two pairs of views are created for measuring the angular offset $\theta$, while using all views to predict the maximal explorable distance $d$. As mentioned in the \textit{Main Paper} \S\ref{subsec:aux_tasks}, the ground-truth angular offset is defined in range $[-\pi,\pi]$, denoting either clockwise (negative) or counter-clockwise (positive) rotation, whichever is smaller in absolute value. The intuition behind such design is to encourage the model to learn the most efficient rotation between two orientations. As for the exploration distance prediction, we collect the ground-truth distances by asking the agent to move forward 5.0 meters in each view direction with a small step size of 0.10 meters and use the \texttt{sim.previous\_step\_collided()} to detect collision.

\subsection{Path Sampling}

The path for connecting two distant views is sampled to create the top-down semantic map for Ego$^2$-Map contrastive learning. Specifically, we consider each viewpoint as a source $p_{s}$ and randomly select a target viewpoint $p_{t}$ within 7.0 meters geodesic distance radius. As described in \textit{Main Paper} \S\ref{subsec:ego2map}, each viewpoint in Ego$^2$-Map is represented by a single egocentric view, which we directly use one of the four sampled images for this purpose. Then, we apply the \texttt{ShortestPathFollower($p_{s}$,$\theta_{s}$,$p_{t}$)} to compute the shortest path (and actions) for traveling from $p_{s}$ to $p_{t}$\footnote{More precisely, from the source view $I_{s}$ to $p_{t}$ since the function does not specify a target orientation.}, where $\theta_{s}$ is the agent's orientation at the source viewpoint. Our agent uses a small turning angle of 5$^{\circ}$ and a small forward step of 0.10 meters, so that more fine-grained and accurate paths can be created. The shortest path is found if the agent's final position is within 0.50 meters of geodesic distance to the target and the total number of actions is less than 140 steps. Otherwise, a new target viewpoint will be paired for examination. Note that we control each view to be either a unique source or target view in all trajectories to avoid repetitive use of the same image in Ego$^2$-Map. Finally, \texttt{ShortestPathFollower()} returns the agent's position, orientation, and action at each time step, which are applied to generate the semantic map.

\subsection{Generating Semantic Maps}

We apply the Semantic MapNet (SMNet)~\cite{cartillier2021semmap} to generate top-down semantic maps from the sampled paths. Briefly, SMNet takes a sequence of the agent's egocentric RGBD observations and poses as input, for each step, the network encodes the RGBD frame and projects it to a floor plan. Then, a spatial memory tensor accumulates the projected egocentric features from all steps. Finally, a map decoder produces the allocentric semantic map from the aggregated memory tensor. We refer the readers to the paper and code\footnote{SMNet: \href{https://github.com/vincentcartillier/Semantic-MapNet}{https://github.com/vincentcartillier/Semantic-MapNet}.} of the SMNet for more details.

In this work, we sightly shift the map so that the agent's starting position $p_{s}$ is at the center of the map. We also scale the map such that it covers a $[-6,6]$ meters range (slightly less than the sampling radius of $p_{t}$). To avoid some rare cases where little semantic information is captured at the target position, we augment the paths by adding a 360$^{\circ}$ rotation at $p_{t}$. Moreover, we highlight the agent's transition by drawing out the path as a gradient color line on the map. Note that the pre-trained SMNet applies a different resolution and field of view for the egocentric images than the views in our image sampling process, we follow SMNet's default configurations to sample the RGBD images on the path.

\begin{figure*}[t]
  \centering
  \includegraphics[width=\textwidth]{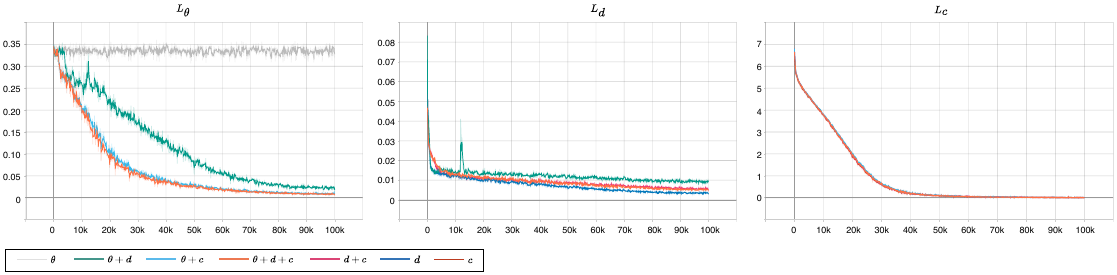}
  \caption{Loss curves of the pre-training objectives. 
  }
  \label{fig:loss_curve}
\end{figure*}

\begin{table*}[t]
  \begin{center}
  \resizebox{0.73\textwidth}{!}{
  \begin{tabular}{l|ccc|c|c|cc}
    \hline \hline
    \multicolumn{1}{c|}{\multirow{2}{*}{Model \#}} & \multicolumn{3}{c}{Pre-Training Objectives} & \multicolumn{4}{|c}{Error / Accuracy} \Tstrut\\
    \cline{2-8} &
    \multicolumn{1}{c}{Angular} & \multicolumn{1}{c}{Explorable} & \multicolumn{1}{c}{Contrastive} & \multicolumn{1}{|c}{$\Delta\theta$ (rad)$\downarrow$} & \multicolumn{1}{|c}{$\Delta{d}$ (m)$\downarrow$} &
    \multicolumn{1}{|c}{I$\rightarrow$M(\%)$\uparrow$} & \multicolumn{1}{c}{M$\rightarrow$I(\%)$\uparrow$}  \Tstrut\\
    \hline \hline
    1 & \checkmark &  &  & 1.564 & -- & -- & -- \\
    2 &  & \checkmark &  & -- & 0.198 & -- & -- \\
    3 &  &  & \checkmark & -- & -- & 92.01 & 92.12 \\
    \hline
    4 &  & \checkmark & \checkmark & -- & 0.239 & 91.79 & 91.70 \\
    5 & \checkmark &  & \checkmark & 0.810 & -- & 91.53 & 91.63 \\
    6 & \checkmark & \checkmark &  & 0.804 & 0.262 & -- & -- \\
    \hline
    7 & \checkmark & \checkmark & \checkmark & 0.819 & 0.256 & 92.02 & 92.03 \\
    \hline \hline
  \end{tabular}}
\end{center}
\vspace{-10pt}
\caption{Validation of the pre-training tasks. $\Delta{\theta}$ and $\Delta{d}$ denote the averaged discrepancy between the predicted angular offset / explorable distance and their ground-truths, I$\rightarrow$M and M$\rightarrow$I are the alignment accuracy from views-pair to maps and from map to views-pairs (evaluated with batch size 128 on 10,000 novel $(I_{s},I_{t},M)$ triplets). \textit{Models} correspond to Table~\ref{tab:abl_obj} in the \textit{Main Paper}.}
\label{tab:abl_acc}
\end{table*}

\subsection{Creating Dataset}
By following the aforementioned procedure, 252,537 viewpoints are sampled from the 800 training environments, from which 500,000 $(I_{\theta_{0}},I_{\theta_{1}})$ views pairs as well as 500,000 $(I_{s},I_{t},M)$ triplets are created for learning $\mathcal{L}_{\theta}$ and $\mathcal{L}_{c}$, respectively. Figure~\ref{fig:data_stats} shows the distribution of the sampled angular offsets (left), explorable distances (middle), and the distance between source and target viewpoints (right). Note that the maximal geodesic distance for sampling ($I_{s},I_{t}$) is 7.0 meters, but the graph displays the actual path length returned by the \texttt{ShortestPathFollower()} function, hence a few paths are longer than 7.0 meters. Overall, we can see that the sampled angular offset and the path length are roughly uniform in the sampling range, whereas the majority of the explorable distances are in the $[0.2,2.0]$ meters range due to the structure of indoor spaces. 

We employ the WebDataset library\footnote{WebDataset: \href{https://github.com/webdataset/webdataset}{https://github.com/webdataset/webdataset}.} for efficient storing and loading data. Specifically, we create 1,000 shards, each containing 500 data points for training, where each data point includes a $(I_{\theta_{0}},I_{\theta_{1}})$ views pair and a $(I_{s},I_{t},M)$ triplet. Following the same procedure, we also create 20 shards for validating the pre-training objectives (see Appendix \S\ref{appendix:B}).

\section{Pre-Training Statistics and Results}
\label{appendix:B}

We present the loss curves and the validation results of the pre-training variants corresponding to Table~\ref{tab:abl_obj} in the \textit{Main Paper}. As shown in Appendix Figure~\ref{fig:loss_curve} and Table~\ref{tab:abl_acc}, the loss $\mathcal{L}_{\theta}$ of the angular offset prediction does not converge to the same level as the others when minimized alone, showing large error in the prediction and leading to invalid features for the downstream navigation tasks. However, it is interesting to see that $\mathcal{L}_{\theta}$ can be learned in the presence of $\mathcal{L}_{d}$ or $\mathcal{L}_{c}$. One possible reason is that the model can extract valuable spatial information by predicting distance or aligning views and maps to facilitate the learning of angular relationships between views. In Model\#5, Model\#6, and Model\#7, the predicted angular error is around 0.810 rad (46$^{\circ}$), again suggesting the difficulty of learning angular offset from views. Besides, we can see that the learning of $\mathcal{L}_{d}$ only shows a minor difference across the model variants, and the prediction error is very low (up to 26.2 centimeters), which indicates that the learned features contain accurate information of whether a direction is explorable. As for the Ego$^2$-Map contrastive learning, the loss curves in training smoothly converge to zero. We evaluate the alignment accuracy with a batch size of 128 on 10,000 novel $(I_{s},I_{t},M)$ triplets from the validation split; results show that both the views-pair to maps (I$\rightarrow$M) and map to views-pairs (M$\rightarrow$I) matching are highly accurate, suggesting the transfer of map information to egocentric representations.

\end{document}